\documentclass[runningheads]{llncs}
\usepackage{graphicx}
\usepackage{amsmath,amsfonts}
\usepackage{amssymb}
\usepackage{pifont}
\usepackage{algorithmic}
\usepackage{multirow}
\usepackage{algorithm}
\usepackage{array}
\usepackage[caption=false,font=normalsize,labelfont=sf,textfont=sf]{subfig}
\usepackage{textcomp}
\usepackage{stfloats}
\usepackage{url}
\usepackage{svg}
\usepackage{verbatim}
\usepackage{orcidlink}
\usepackage{cite}

\newcommand{\lsade}{\texttt{LSADE}}
\newcommand{\esa}{\texttt{ESA}}
\newcommand{\gors}{\texttt{GORS-SSLPSO}}
\newcommand{\msmto}{\texttt{MS-MTO}}
\newcommand{\tlssl}{\texttt{TL-SSLPSO}}
\newcommand{\bis}{\texttt{BiS-SAHA}}
\newcommand{\ikaea}{\texttt{IKAEA}}
\newcommand{\samso}{\texttt{SAMSO}}
\newcommand{\tsdd}{\texttt{TS-DDEO}}
\newcommand{\rs}{\texttt{RS}}
\newcommand{\shpso}{\texttt{SHPSO}}
\newcommand{\ssla}{\texttt{SSL-A-PSO}}

\newcommand{\xmark}{\ding{55}}%

\begin{document}
\title{Performance Comparison of Surrogate-Assisted Evolutionary Algorithms on Computational Fluid Dynamics Problems\thanks{This work was supported by the project IGA BUT No. FSI-S-23-8394 ``Artificial intelligence methods in engineering tasks'' and by the project (no. 22-31173S) ``Benchmarking derivative-free global optimization methods'' funded by the {Czech Science Foundation}.}}
%
%
\author{Jakub K\r{u}dela\orcidlink{0000-0002-4372-2105} \and
Ladislav Dobrovsk\'{y}\orcidlink{0000-0002-7186-7213} 
}
\authorrunning{J. K\r{u}dela and L. Dobrovsk\'{y}}
%
\institute{Institute of Automation and Computer Science, Faculty of Mechanical Engineering, Brno University of Technology\\
\email{jakub.kudela@vutbr.cz}
}
\maketitle              
\begin{abstract}
Surrogate-assisted evolutionary algorithms (SAEAs) are recently among the most widely studied methods for their capability to solve expensive real-world optimization problems. However, the development of new methods and benchmarking with other techniques still relies almost exclusively on artificially created problems. In this paper, we use two real-world computational fluid dynamics problems to compare the performance of eleven state-of-the-art single-objective SAEAs. We analyze the performance by investigating the quality and robustness of the obtained solutions and the convergence properties of the selected methods. Our findings suggest that the more recently published methods, as well as the techniques that utilize differential evolution as one of their optimization mechanisms, perform significantly better than the other considered methods.

\end{abstract}
\section{Introduction}
The field of evolutionary computation (EC) has produced a multitude of pivotal evolutionary (or metaheuristic) algorithms (EAs) \cite{kudela2022critical}, such as genetic algorithm (GA) \cite{mitchell1998introduction}, evolutionary strategy (ES) \cite{beyer2002evolution}, differential evolution (DE) \cite{storn1997differential}, or particle swarm optimization (PSO) \cite{kennedy1995particle}, that were used to solve a wide range of different optimization problems \cite{bujok2023differential,gong2021nonlinear}. However, these standard EAs usually require a large number of objective/constraint function evaluations to find satisfactory solutions, which severely limits their utility for solving expensive real-world problems \cite{jin2018data}. In applications such as aerodynamic design, computational fluid dynamics (CFD), or finite element method analysis, a single function evaluation requires running expensive computations that may take several minutes to several hours \cite{kudela2022recent}. The need for such high-fidelity simulations imposes a practical limit on the number of designs that may be considered during optimization.

The interest in computationally expensive optimization has increased substantially in the past few years \cite{kudela2022recent}. Perhaps the first widely known example of such problems was presented in \cite{jones1998efficient} where the two applications were in integrated circuit design and in the automotive industry. Since then, there have been numerous applications of optimization for such expensive problems, especially in the field of CFD. A multi-objective shape optimization of aerofoils (accounting for low-drag and high-lift) was investigated in \cite{naujoks2002evaluating}. A CFD-based shape optimization problem that aimed to minimize the mass of beams under structural constraints was presented in \cite{leary2004derivative}. In a heat exchanger design, a multi-objective formulation (that maximized heat flux and minimized pressure drop) was proposed in \cite{foli2006optimization}. More recent applications are concerned with CFD-based geometry optimization problems, such as the minimization of pressure differences in pipes \cite{daniels2016shape} and ducts \cite{daniels2017automatic}, and in ship hydrodynamic optimization \cite{serani2016ship}, or in multiobjective ship hull design \cite{campana2018multi}.

One of the approaches for solving these computationally expensive problems is to construct surrogate models (also called meta-models) to assist the EAs. Such surrogate models are generally constructed using historical data to approximate and predict the landscape of the objective function, with a negligible computational cost. These types of algorithms are called data-driven EAs \cite{jin2018data} or surrogate-assisted EAs (SAEAs) \cite{liu2013gaussian}. Depending on whether real function evaluations can be performed during the optimization process, SAEAs can be further divided into two categories: offline SAEAs and online SAEAs \cite{jin2018data}. The offline SAEAs prioritize building the most suitable surrogate model based on available historical data to predict the position of the optimal solution \cite{wang2018offline}. In contrast to the offline SAEAs, the online SAEAs work by effectively sampling candidate solutions that will be evaluated on the real expensive function, and then update the corresponding surrogate models and populations in the process of optimization \cite{jin2018data}.

The majority of SAEAs are constructed on the basis of standard EAs and are organically combined with surrogate models for predicting expensive real function evaluations \cite{zhen2022evolutionary}. Various EAs, such as GA \cite{zhou2006combining}, DE \cite{liu2013gaussian}, or PSO \cite{yu2018surrogate} have been successfully used in SAEAs and, in recent years, there has been a multitude of SAEAs proposed in the literature. However, to develop these methods and compare them with other SAEAs, most authors rely almost exclusively on artificially created problems. 

As EC methods in general (as SAEAs in particular) are difficult to analyze analytically, the majority of the reasoning about their utility is done by benchmarking \cite{hellwig2019benchmarking}. Throughout the years, many different benchmark sets and functions were introduced in journal articles \cite{garcia2017since,kudela2022new}, but the most widely-used ones were developed for competitions (and special sessions) on black-box optimization at two high-profile conferences, the Genetic and Evolutionary Computation Conference (GECCO), and the IEEE Congress on Evolutionary Computation (CEC). However, the use of these benchmark sets is not without critique, with some authors voicing their concern about the artificial nature of these problems \cite{piotrowski2015regarding} and advocating for benchmarking EAs on real-world problems instead \cite{tzanetos2021nature}. Notable exceptions are the recent works on comparing low-budget methods on the OpenAI Gym \cite{raponi2023optimizing} and on the EXPObench library \cite{bliek2023benchmarking}. 

Although there are quite a lot of applications of SAEAs, the authors of these applications generally develop custom codes for the problems. They are also often reluctant to release the source codes, as many of these applications are proprietary in nature \cite{daniels2018suite}. This means that despite having numerous published successful applications of SAEAs, it is generally difficult to use the simulators for the benchmarking of different methods. The SAEA community is now actively looking for real-world computationally expensive problems that can serve as benchmarks. The vast majority of published SAEAs algorithms that are not tied to a specific application are benchmarked on a handful of problems (such as the Ellipsoid, Rosenbrock, Ackley, and Griewank functions) from the CEC competitions \cite{liang2013problem,chen2014problem}. However, these are only pseudo-expensive problems, i.e. problems that are inexpensive in nature and use artificial delays or restrictions on the number of function evaluations to mimic the expensive problems \cite{daniels2018suite}.

This lack in the availability of real-world-based benchmark sets of expensive problems for SAEAs spurred the creation of the CFD suite presented first at the PPSN XV conference \cite{daniels2018suite}. The problems in this suite focus on designing different mechanisms, using CFD to evaluate the performances of the candidate geometries in a fluid environment. Other places to find computationally expensive problems that are well-suited for benchmarking are the GECCO Industrial Challenge Competitions, with one of them focusing on CFD problems \cite{gecco2020}, while others aimed at a hospital planning application \cite{gecco2021,bartz2020hospital}. These industrial problems were used to benchmark parallel SAEA methods in \cite{rehbach2018comparison}, but a thorough computational comparison of the state-of-the-art SAEAs on such problems was still missing.

In this paper, we aim to fill this research gap by benchmarking state-of-the-art single-objective SAEAs on the CFD problems from the suite \cite{daniels2018suite}. As the representative methods, we selected eleven SAEAs that were recently published either in high-profile journals or conferences and were developed on similar test problems (in terms of dimensions and available function evaluations). We analyze the performance of the selected SAEAs by investigating the quality and robustness of the best-found solutions and their convergence properties. In the analysis, we follow the recently published guidelines for comparing EAs \cite{latorre2021prescription}. The codes for all the methods used in the computational comparison, and the code used to run the CFD experiments are made publicly available in the Zenodo repository \cite{zenodolink}.

The rest of this paper is organized as follows. Section \ref{s2} introduces the framework of surrogate-assisted optimization. In Section \ref{s3} we introduce the two CFD problems used for the computational comparisons. Section \ref{s4} contains a brief discussion of the selected state-of-the-art SAEAs. Section \ref{s5} reports on the results of the computational comparisons. Section \ref{s6} concludes the paper and discusses implications and future work.

\begin{algorithm}[H]
\caption{Prototypical structure of a surrogated-assisted optimization method \cite{karlsson2020continuous}.}\label{alg:alg1}
\begin{algorithmic}
\STATE 
\STATE {\textbf{Require:}} budget $B$, surrogate model $M$, acquisition function $A$
\STATE \hspace{0.5cm} Initialize $x^{(1)}$ randomly and an empty set $H$
\STATE \hspace{0.5cm} \textbf{for } $m=1:B$ \textbf{do}
\STATE \hspace{1cm} $y^{(m)}\leftarrow f(x^{(m)})$
\STATE \hspace{1cm} $H\leftarrow H \cup \{ (x^{(m)},y^{(m)}) \}$
\STATE \hspace{1cm} $M\leftarrow $ fit surrogate model using $H$
\STATE \hspace{1cm} $x^{(m+1)}\leftarrow \text{argmax}_x A(M,x)$
\STATE \hspace{0.5cm} \textbf{end for}
\STATE \hspace{0.5cm} \textbf{return} best found $(x^*,y^*) \in H$
\end{algorithmic}
\label{alg1}
\end{algorithm}

\section{Surrogate-Assisted Optimization}\label{s2}
In this paper, we consider the class of so-called black-box optimization problems, where the objective function $f:\mathbb{R}^D \rightarrow \mathbb{R}$ has no closed-form expression and only obtainable information about $f$ come from observing its output when evaluating $f(x)$ given some input (decision variable) $x \in \mathbb{R}^D$. The prototypical box-constrained black-box optimization problem has the following form \vspace{-2mm}
\begin{align} \nonumber
    \text{minimize} \quad &  f(x) \\ \label{eq1}
    \text{subject to} \quad & x \in \mathbb{R}^D \\ \nonumber
    & l_i \leq x_i \leq u_i, \quad i=1,\dots,D,
\end{align}
where $l_i, u_i$ are the lower and upper bounds on the decision variable $x_i$, respectively. The evaluation of $y = f(x)$ is assumed to be computationally expensive (i.e., requiring running time-consuming CFD simulations). This implies that we are mostly interested in finding good (and not necessarily optimal) solutions to (\ref{eq1}) in a reasonable amount of time. We approach this by setting a limited budget $B$ for the number of available calls of $f$.

One of the possible approaches for this class of problems is to utilize a so-called surrogate model, which can be thought of as an auxiliary function $M$ that we use to approximate the objective function $f$. The surrogate model should be cheaper to evaluate
than the original black-box function $f$ and is usually constructed using the evaluation history ($m$ already evaluated points) $H = \{ (x^{(1)}, y^{(1)}),\dots, (x^{(m)}, y^{(m)}) \}$. Kriging (also known as Gaussian Process) models and Radial Basis Functions (RBFs) are among the most widely used approaches for generating surrogate models \cite{kudela2022recent}. Other techniques that are also used with SAEAs are polynomial response surface methods \cite{krithikaa2016differential} and support vector machines \cite{loshchilov2010comparison}, with various different techniques also considered \cite{luo2020novel,zhou2020neighborhood,bliek2021black}. It was found that the Kriging models outperform other surrogates in solving low-dimensional optimization problems, while RBFs are the more efficient surrogates for solving high-dimensional optimization problems \cite{diaz2017comparison}. 

The main purpose of the surrogate model is to predict the next promising points for evaluation, which is typically also guided by an acquisition function $A(M,x)$. This function predicts how promising a new point $x$ is, balancing the trade-off of exploration (the search in regions where the surrogate model displays high levels of uncertainty) and exploitation (the search near already evaluated points that resulted in low objective values). The prototypical structure of a surrogate-assisted optimization method is shown in Algorithm \ref{alg:alg1}. Most SAEAs initialize the population by evaluating a fixed number of points selected by the Latin Hypercube Sampling method \cite{helton2003latin}.

\section{Selected Problems}\label{s3}
As the performance of a given engineering design in a fluid environment usually cannot be evaluated analytically, we often resort to various numerical CDF approximations. These CFD computations require the solution of a set of Partial Differential
Equations (PDEs) describing the physics of the flow of the fluid, which is typically approached using finite volume (and similar) methods. There are numerous software products available that can perform these calculations. One of the most used open-source 
options is the C++ code OpenFOAM \cite{weller1998tensorial}. Both of the selected problems, described in greater detail in \cite{daniels2018suite} and \cite{rehbach2018comparison}, use OpenFOAM for the CFD simulation. The problem geometries generated from the optimization variables are transformed into sterolithography (STL) files and imported into this OpenFOAM framework \cite{daniels2019review}. Similar setup was recently used in \cite{bliek2023benchmarking}, but with a completely different set of algorithms used for the comparison.

\subsection{PitzDaily}
The separation of flows, reattachment, and recirculation are all phenomena commonly observed in various engineering applications. Such features are usually undesirable \cite{daniels2018suite}. The first CFD problem is based on an experimental study by Pitz and Daily \cite{pitzdaily1981} (hence the name), which features a so-called ``backward-facing step'', that is often used as a simplified prototype for studying the flow phenomena mentioned above. In such a geometry (which can be found in Fig. \ref{fig:curves} under the ``PitzDaily boundary'' label), the flow separates at the edge of the step and creates a recirculation zone. Afterward, the flow reattaches at a distance beyond this step. This case study has been used as a benchmark case for the accuracy testing of different CFD methodologies.

One of the undesirable characteristics of a flow is head losses. We can describe this quantity (the mechanical energy loss factor $\zeta$) as the energy that is transformed to a form which can no longer be used in the operation of an energy-producing, conducting, or consuming system (for instance because of the frictional losses, or dissipation due to turbulence) \cite{daniels2018suite}. The energy loss factor $\zeta$ of a given design can be defined as the total pressure difference between the inflow and outflow of the mechanism (taken relative to the kinetic energy at the inflow) 
\begin{equation*}
    \resizebox{\hsize}{!}{%
    $     \zeta = \frac{2}{\rho U^2_{\text{in}}} \left[ \frac{1}{A_{\text{in}}} \int_{\text{in}} P_{t,\text{in}}(u \cdot n) dA_{\text{in}} -  \frac{1}{A_{\text{out}}} \int_{\text{out}} P_{t,\text{out}}(u \cdot n) dA_{\text{out}} \right]     $
    }
\end{equation*}
where $\rho$ is the density of the fluid, $U_\text{in}$ is the inflow velocity, $A$ is the cross-sectional area, $P_t$ is the total pressure, $u \cdot n$ describes the velocity component normal to the boundary, and subscripts ``in'' and ``out'' indicate the inflow and outflow boundaries. The primary objective of this problem is to minimize the energy loss $\zeta$ by changing the geometry of the design.

The standard procedure to create such geometry would be to use some Computer-Aided Design (CAD) software. The problem is that automatically altering designs using CAD is very challenging. Therefore, we generally resort to different parametric representations of the parts of the original CAD geometry. We can then generate new geometries by changing the parameters of such representation. 

\begin{figure}[!t]
\centering
\includegraphics[width=.6\linewidth]{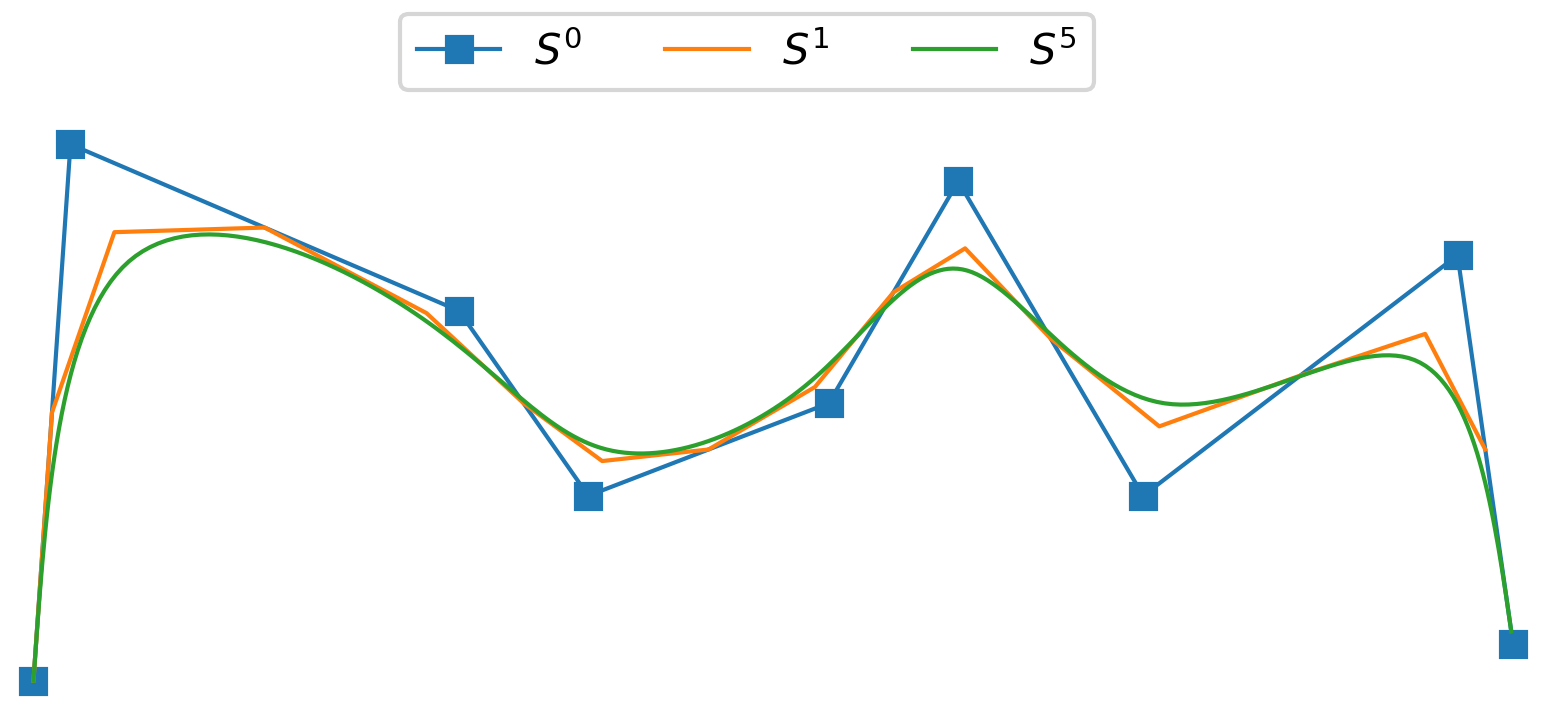}
\caption{Example of the Catmull-Clark subdivision curve. $S^0$ are the original control points, $S^1$ is the approximation of $\mathcal{C}$ after one iteration, $S^5$ is the visually smooth curve after five iterations.}
\label{fig:catmull}
\end{figure}

For the PitzDaily problem, we use the Catmull-Clark subdivision curves \cite{catmull1998recursively} to alter the boundary wall. In this method, the curve $\mathcal{C}$ is parameterized by a sequence of $n$ vertices (or control points) $S^0 = \langle v_1, \dots , v_n\rangle $. In each iteration, a mid-point between adjacent vertices is inserted, and the positions of the vertices are adjusted, generating a larger sequence. From the practical perspective, only a few iterations of these subdivisions are necessary for a visually smooth curve (see Fig. \ref{fig:catmull}) exportable in the STL format. In this paper, the iteration limit was set to five. We chose the positions of five control points of the Catmull-Clark curve in 2D as the decision vector (see Fig. \ref{fig:curves}), resulting in an optimization problem with 10 variables.

\subsection{Electrostatic Precipitator}
\begin{figure}[!t]
\centering
\includegraphics[width=.65\linewidth]{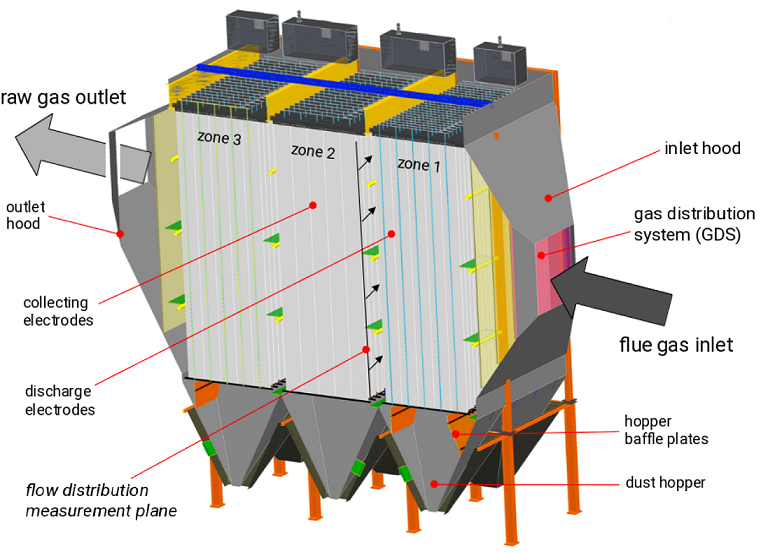} \hspace{2mm} \includegraphics[width=.2\linewidth]{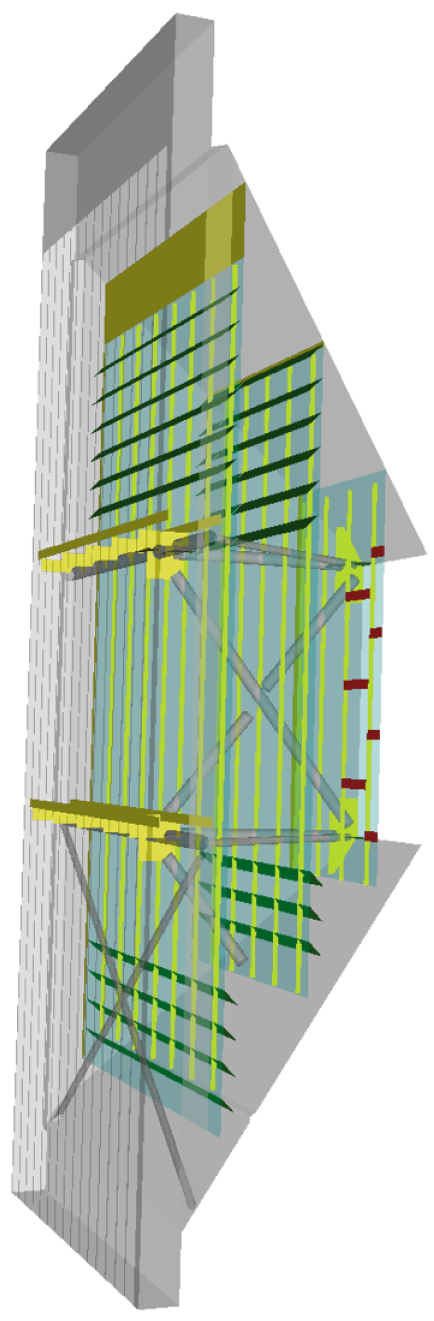}
\caption{ESP with 3 separation zones (left) and GDS mounted in the inlet hood of an ESP (right) \cite{gecco2020}.}
\label{fig_ESP}
\end{figure}

The Electrostatic Precipitator (ESP) is a real-world problem from industrial optimization that was first proposed in \cite{rehbach2018comparison}. The ESP is one of the crucial components in gas cleaning systems used in combustion power plants (or similar industries) to remove solid particles from gas streams (pollution reduction). Fig. \ref{fig_ESP} shows a schema of such a system.

To control the flow of the gas flow through the different separation zones (where the removal of particles from the exhaust gases occurs) a gas distribution system (GDS) is needed. If there were no GDS used (or the system was not well configured), the fast inlet gas stream would run through the separation zones, resulting in low separation efficiencies. For the efficient operation of the EPS, the GDS must have a good configuration. In our case, the GDS has 49 slots, which can be configured with baffles (metal plates mounted at an angle to the flow of the gas), blocking plates (block the gas stream), or perforated plates (slow down and partially block the gas stream).

In total, there are 8 different options available for each of the 49 configurable slots (decision variables), resulting in a solution space of roughly $10^{44}$ possibilities. Even though this problem is discrete in nature, it has been shown in \cite{karlsson2020continuous} that continuous SAEAs are well-suited to solve it. The ESP problem was also the focus of the GECCO 2020 Industrial Challenge \cite{gecco2020}.

\section{Selected Methods and Experimental Setup}\label{s4}
For the selection and comparison of representative SAEAs, we followed the guidelines published in \cite{latorre2021prescription}. The selected methods were all recently developed algorithms that were published either in high-profile journals (such as IEEE Transactions on Cybernetics, IEEE Transactions on Evolutionary Computation, Information Sciences, etc.) or conferences (such as GECCO). All of the methods were also trained on similar dimensions (between 10 and 100, which covers the range of our two CFD problems), using a similar computational budget.

\begin{table}[]
\caption{SAEAs selected for computational comparison. The ``Real?'' column shows if the method was evaluated on real-world problems.}
\label{tab:algos}
\centering
\resizebox{0.7\linewidth}{!}{
\begin{tabular}{llllll} \hline 
\multirow{2}{*}{Method name} & \multirow{2}{*}{Year} & \multirow{2}{*}{Ref.} & Surrogate & Optimization & \multirow{2}{*}{Real?} \\
& & & model & method \\ \hline 
\shpso       & 2018 &  \cite{yu2018surrogate}    & RBF             & PSO                 & N     \\
\ssla   & 2018 & \cite{sun2018semi}     & RBF             & PSO                 & N     \\
\gors & 2019 &  \cite{yu2019generation}    & RBF             & PSO                 & N     \\
\msmto      & 2020 & \cite{liao2020multi}     & RBF             & MTO                  & N     \\
\tlssl   & 2020 & \cite{yu2020truncation}     & RBF             & PSO                 & Y     \\
\bis    & 2021 &  \cite{ren2021bi}    & RBF             & PSO, DE                  & Y     \\
\ikaea       & 2021 &  \cite{zhan2021fast}    & Kriging         & DE                  & N     \\
\samso       & 2021 &  \cite{li2020surrogate}    & RBF             & TLBO, PSO           & N     \\
\tsdd     & 2021 &  \cite{zhen2021two}    & RBF             & PSO, DE             & N     \\
\esa         & 2022 &  \cite{zhen2022evolutionary}    & RBF             & DE                  & N     \\
\lsade       & 2022 & \cite{kuudela2023combining}     & RBF, Lipschitz  & DE, SQP             & N    \\ \hline 
\end{tabular}
}
\end{table}

Table \ref{tab:algos} shows a brief overview of the SAEAs selected for the computational comparison in chronological order. We can find that the most used surrogate model was RBF, and the most utilized EAs were PSO and DE. Only two of the selected methods were tested on real-world problems, with the rest relying purely on artificially created ones. However, these two real-world problems were not computationally expensive (both used analytical expressions in the formulations).

Another common feature of all of the selected methods was that they had their respective codes publicly available, making it possible for verification of the presented results as well as for conducting additional computational comparisons. The implementation of the selected SAEAs, as well as the important information about their parametrization, and the implementation of the CFD problems, can be found in a public Zenodo repository \cite{zenodolink}.

\subsection{Experimental Setup}
The selected SAEAs were implemented in MATLAB R2022b and the experiments were run on a workstation with 3.7 GHz AMD Ryzen 9 5900X 12-Core processor and 64 GB RAM. As the selected SAEAs are stochastic methods, each of them was run 24 times (efficiently using the 12-core machine) on the two CFD problems, in order to get statistically representative results and provide a solid basis for algorithmic comparison. To give context to the results, we also implemented a simple random search (\rs) routine. The function evaluation budget was set to $B = 1{,}000$ in both cases. For both problems, the time needed for the evaluation of the objective function (i.e., running the CFD simulations) was approximately 30\,s.

\begin{table}[]
\caption{Statistics of the 24 runs of the selected SAEAs on the PitzDaily model. The best three methods in each category are highlighted in bold.}
\label{tab:pitz}
\centering
\resizebox{0.8\linewidth}{!}{
\begin{tabular}{lcccccr} \hline 
Method      & Min      & Mean     & Median   & Max      & Std      & \multicolumn{1}{c}{Rank} \\ \hline 
\rs          & 8.59E-02 & 9.26E-02 & 9.24E-02 & 9.63E-02 & 3.19E-03 & 11.71         \\
\shpso       & 8.03E-02 & 8.31E-02 & 8.34E-02 & 8.61E-02 & 1.62E-03 & 5.50          \\
\ssla   & 8.06E-02 & 8.46E-02 & 8.31E-02 & 9.34E-02 & 3.21E-03 & 5.96          \\
\gors & 8.08E-02 & 8.62E-02 & 8.52E-02 & 9.42E-02 & 3.26E-03 & 8.63          \\
\msmto      & 8.31E-02 & 8.62E-02 & 8.52E-02 & 9.47E-02 & 3.04E-03 & 9.08          \\
\tlssl   & 8.28E-02 & 8.45E-02 & 8.32E-02 & 9.35E-02 & 2.90E-03 & 6.04          \\
\bis    & 8.31E-02 & 8.52E-02 & 8.42E-02 & 9.14E-02 & 2.27E-03 & 8.50          \\
\ikaea       & \textbf{7.95E-02} & \textbf{8.28E-02} & \textbf{8.30E-02} & \textbf{8.88E-02} & 2.31E-03 & \textbf{4.17}          \\
\samso       & 8.00E-02 & 8.38E-02 & 8.33E-02 & 8.99E-02 & 1.79E-03 & 6.17          \\
\tsdd     & 7.98E-02 & 8.29E-02 & \textbf{8.30E-02} & \textbf{8.79E-02} & 1.67E-03 & 4.50          \\
\esa         & \textbf{7.94E-02} & \textbf{8.23E-02} & \textbf{8.27E-02} & \textbf{8.56E-02} & 1.98E-03 & \textbf{4.08}          \\
\lsade       & \textbf{7.95E-02} & \textbf{8.24E-02} & 8.31E-02 & 9.13E-02 & 2.48E-03 & \textbf{3.67}         \\ \hline 
\end{tabular}
}
\end{table}

\section{Results and Discussion}\label{s5}

\subsection{PitzDaily Results}
\begin{figure}[!t]
\centering
\includegraphics[width=0.47\linewidth]{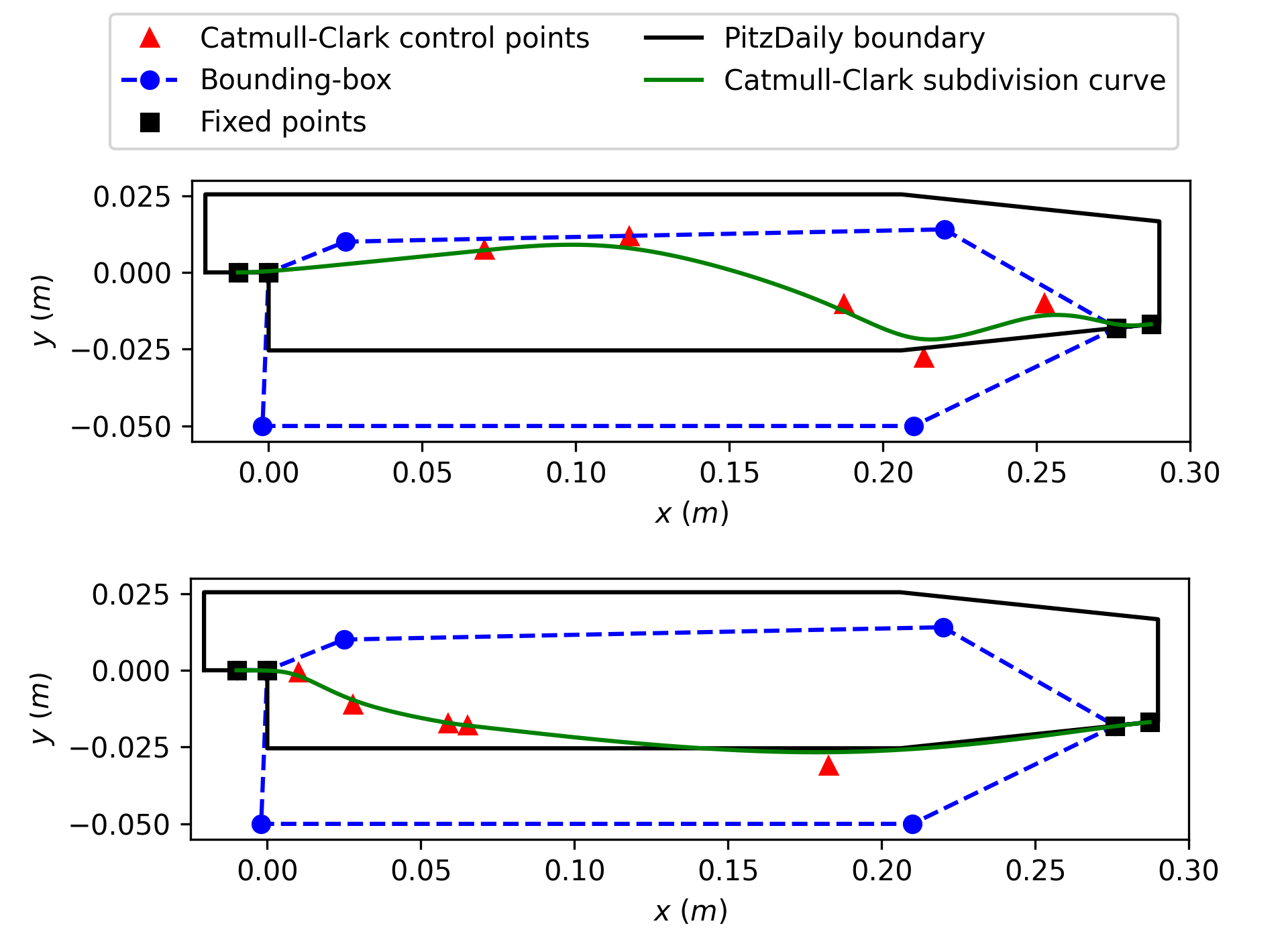} \includegraphics[width=0.47\linewidth]{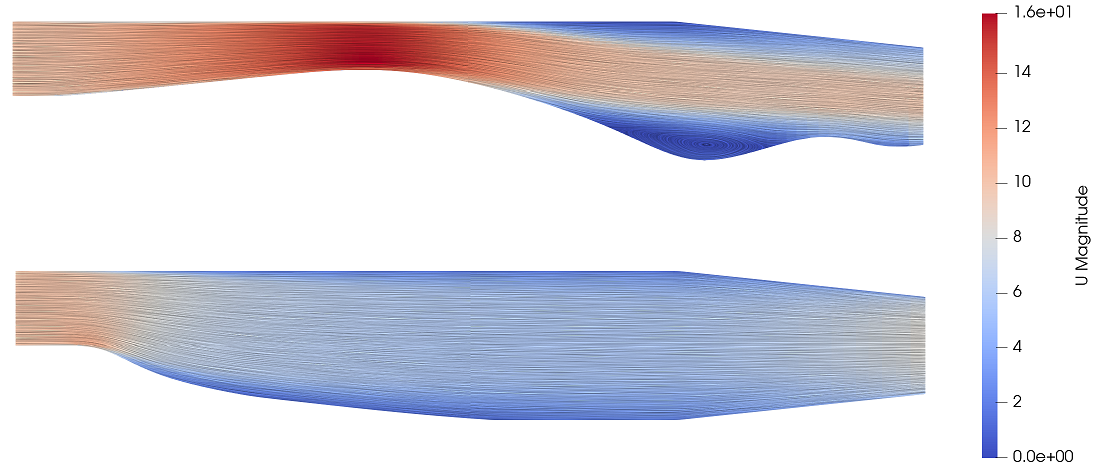}
\caption{Random design (top) and best-found design (bottom) for the PitzDaily problem (left), flowfield (CFD) simulations of these designs (right).}
\label{fig:curves}
\end{figure}

The results of the computations (statistics of the 24 independent runs) on the PitzDaily problem (with $D=10$) are summarized in Table \ref{tab:pitz}. The Catmull-Clark curve of the best solution (i.e., the best design shape) found by any method (in this case, \esa) and a randomly generated solution are depicted in Fig. \ref{fig:curves}, while the flowfield (CFD) simulations for these designs are shown in Fig. \ref{fig:paraview} (these were generated using the ParaView software). On the best-found design, we can observe a much smoother transition of $u$ from one boundary to the other, resulting in a lower value of the energy loss factor $\zeta$.

\begin{figure}[!ht]
  \centering
  \includegraphics[width=0.7\linewidth]{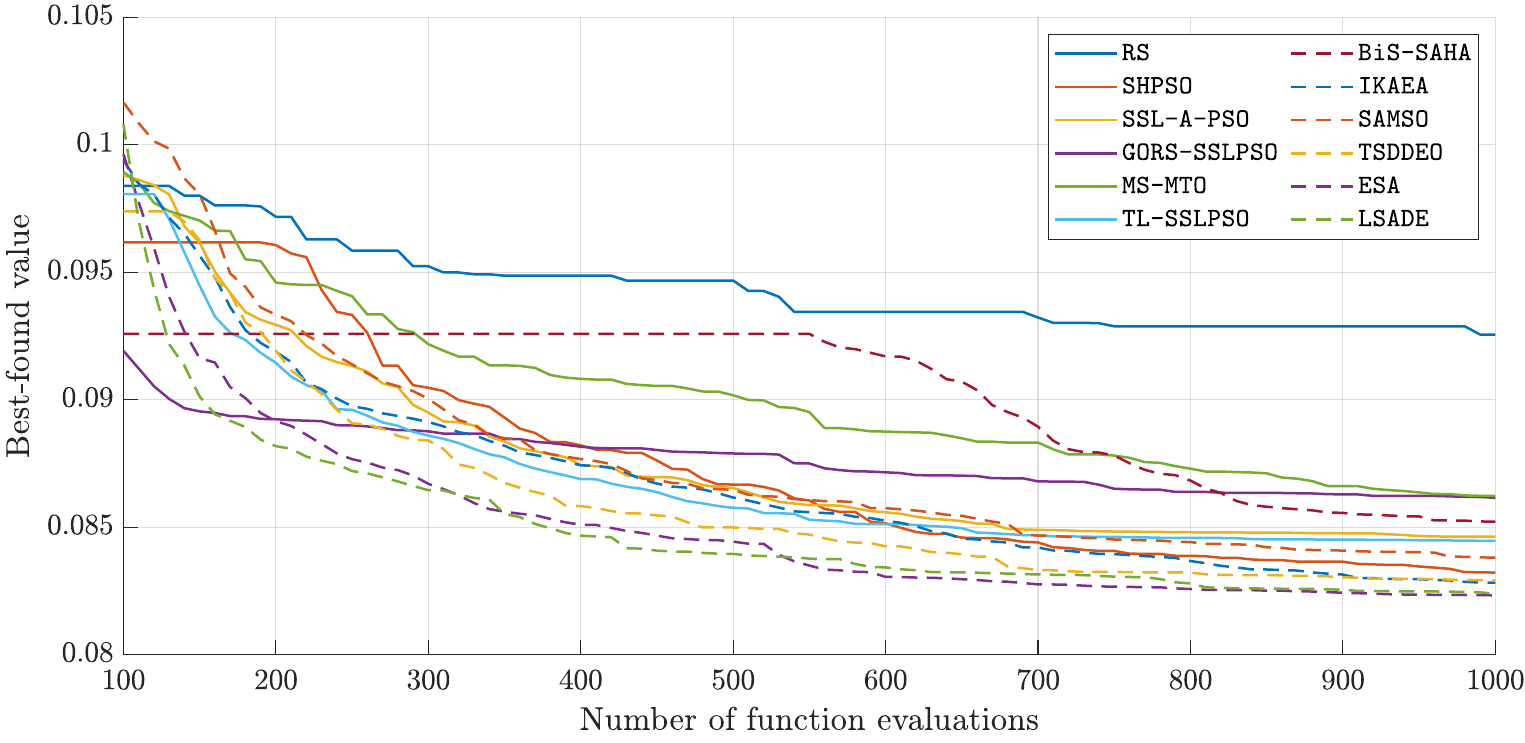} 
  \caption{Convergence of the average best-found value of the selected SAEAs on the PitzDaily problem.}
  \label{fig:conv_pitz}
\end{figure}

\begin{figure}[!ht]
  \centering
  \includegraphics[width=.49\linewidth]{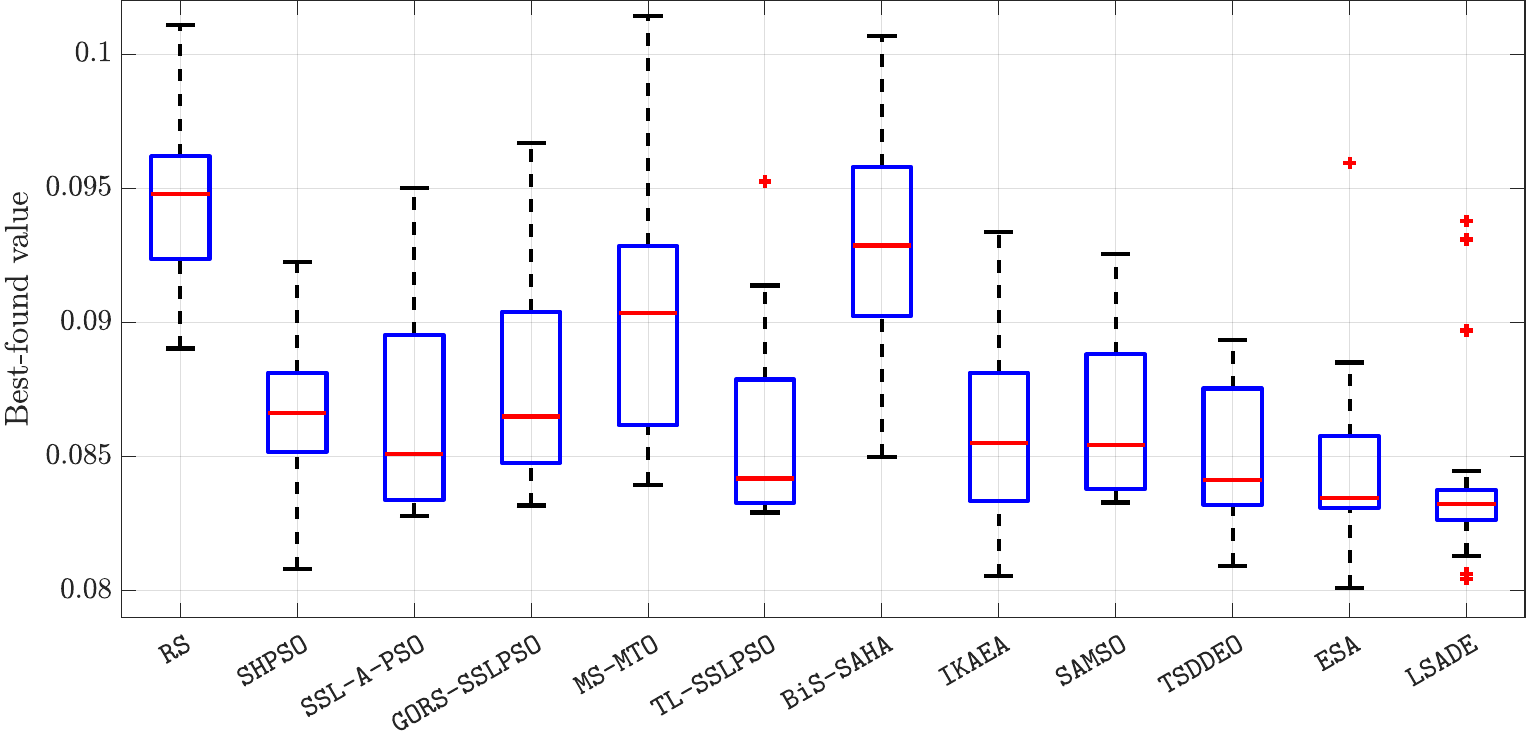} \includegraphics[width=.49\linewidth]{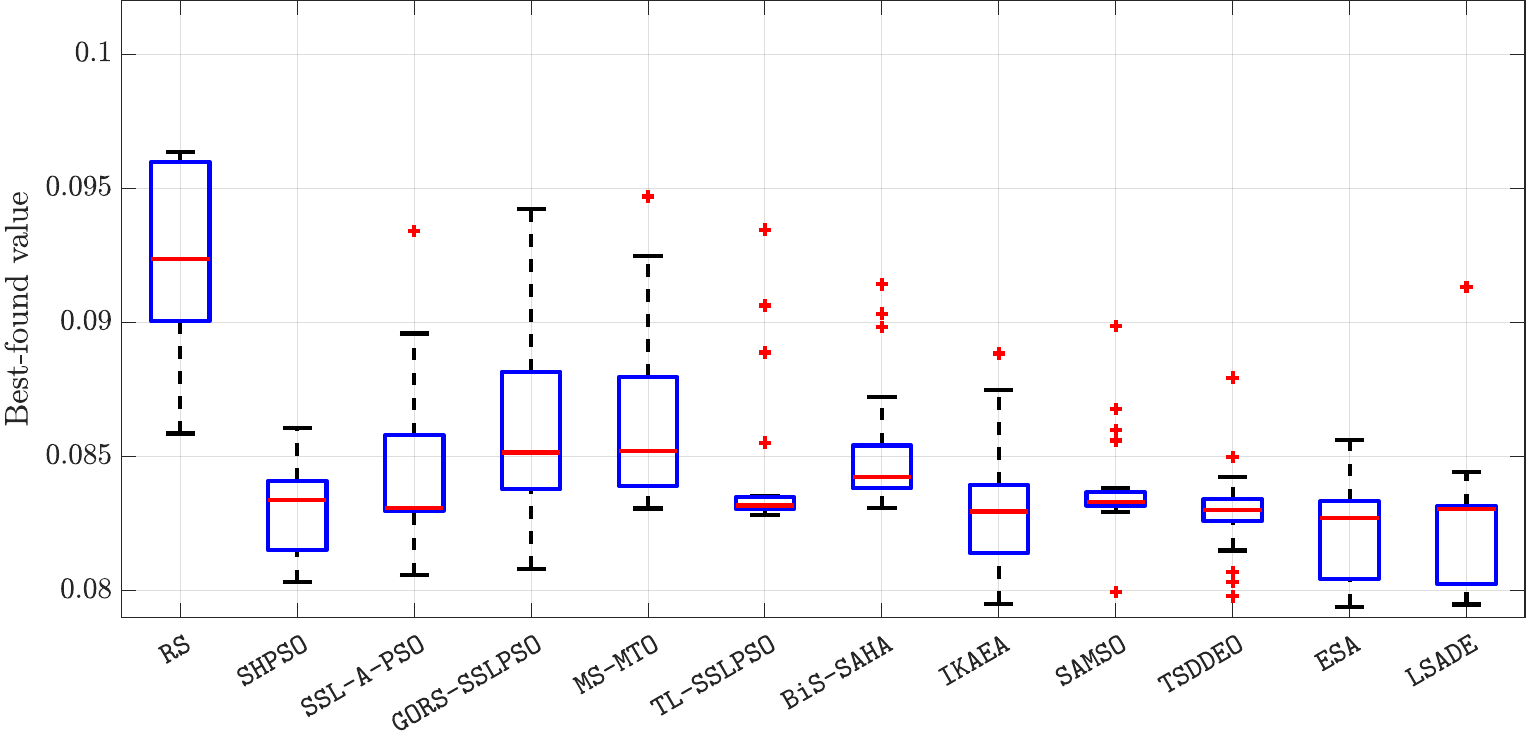}
  \caption{Boxplots of the solutions of the SAEAs on the PitzDaily problem with $B=500$ (top) and $B=1000$ (bottom).}
  \label{fig:box_pitz}
\end{figure}

Based on the statistics shown in Table \ref{tab:pitz}, the best methods for the PitzDaily problem were \esa, \lsade, \ikaea, and \samso. A different insight can be found when looking at the convergence plots of average best-found solutions of the methods shown in Fig. \ref{fig:conv_pitz}. Here, we find that \lsade, \esa, and \tsdd{} were able to find good solutions faster than the other considered SAEAs. These three are among the most recently proposed methods (in Fig. \ref{fig:conv_pitz}, the convergence of methods published after 2021 is shown as a dashed line). As Fig. \ref{fig:conv_pitz} shows only the average performance of the selected SAEAs, we also boxplots of the solutions found in the 24 runs with two different budgets $B = 500$ and $B = 1{,}000$, that are shown in Fig. \ref{fig:box_pitz}. Here, we can find that \lsade{} had a very good performance on both budgets that was somewhat worsened by a single run for which the method was not able to find a good solution. Also, the \tsdd{} method showed only a marginal improvement upon \shpso{} (its base method), which was among the best-performing methods for this problem, despite being the oldest one.

To perform a solid statistical comparison of the selected algorithms on the PitzDaily problem, we followed the guidelines published in \cite{latorre2021prescription}. First, we used the Friedman test to find if significant differences are present among all the algorithms on the $B=500$ and $B=1{,}000$ budgets (the Friedman ranks of the methods for $B=1{,}000$ are shown in Table \ref{tab:pitz}). The p-values for this test were 1.16E-18 (and 1.50E-12 when omitting \rs) for $B=500$ and 2.37E-21 (and 6.46E-13 when omitting \rs) for $B=1{,}000$. As all these p-values values were much lower than the recommended confidence level $\alpha = 0.05$, we can state that there are statistically significant differences between the selected SAEAs. Furthermore, we utilized the Wilcoxon signed-rank test to find if there exist statistically significant differences between the best algorithm (in this case, the \lsade{} algorithm with the lowest Friedman rank) and the other SAEAs. Once the pairwise p-values were obtained, we applied the Holm-Bonferroni \cite{holm1979simple} correction method which counteracts the effect of multiple comparisons by controlling the family-wise error rate \cite{aickin1996adjusting}. The results of the analysis are presented in Table \ref{tab:pitz_p}. For the $B=500$ budget, there were 6 methods that tied with \lsade{} as the best-performing ones. For the $B=1{,}000$ budget, the list of the methods that tied with \lsade{} changed - \tlssl{} and \samso{} dropped, while \shpso{} entered. Forming a union of these two lists, we find that four methods (\ssla, \ikaea, \tsdd, and \esa) were found to be as good as \lsade{} on the PitzDaily problem. Four of these five methods were proposed after 2021, and also four of the five utilize DE in some capacity (in both cases, the exception was \ssla).

\begin{table}[]
\caption{Statistical analysis of the comparison of the selected SAEAs on the PitzDaily problem.}
\label{tab:pitz_p}
\centering
\resizebox{0.7\linewidth}{!}{
\begin{tabular}{lcc|lcc}\hline 
            \multicolumn{3}{c|}{$B = 500$}  &     \multicolumn{3}{c}{$B = 1{,}000$}        \\ 
\lsade{} vs.    & $p$        & $p^*$      & \lsade{} vs.    & $p$        & $p^*$      \\  \hline 
\shpso       & 2.46E-03 & 1.72E-02 & \shpso       & 8.65E-02 & 3.46E-01\xmark \\
\ssla   & 1.91E-02 & 9.57E-02\xmark & \ssla   & 1.10E-02 & 5.50E-02\xmark \\
\gors & 8.05E-05 & 6.44E-04 & \gors & 8.05E-05 & 7.25E-04 \\
\msmto      & 2.67E-05 & 2.67E-04 & \msmto      & 2.35E-05 & 2.35E-04 \\
\tlssl   & 2.78E-02 & 1.11E-01\xmark & \tlssl   & 6.64E-03 & 3.99E-02 \\
\bis    & 5.61E-05 & 5.05E-04 & \bis    & 3.96E-04 & 3.17E-03 \\
\ikaea       & 4.55E-02 & 1.37E-01\xmark & \ikaea       & 3.46E-01 & 8.71E-01\xmark \\
\samso       & 1.10E-02 & 6.60E-02\xmark & \samso       & 2.46E-03 & 1.72E-02 \\
\tsdd     & 1.23E-01 & 2.32E-01\xmark & \tsdd     & 2.90E-01 & 8.71E-01\xmark \\
\esa         & 1.16E-01 & 2.32E-01\xmark & \esa         & 8.19E-01 & 8.19E-01\xmark \\ \hline 
\end{tabular}
}
\flushleft
$p$: p-value computed by the Wilcoxon text \\
$p^*$: p-value corrected with the Holm-Bonferroni procedure \\
\xmark: statistical differences do not exist with significance level $\alpha = 0.05$
\end{table}

\subsection{ESP Results}
The results of the computations on the ESP problem (with $D=49$) are summarized in Table \ref{tab:esp}. Unfortunately, we do not have an expressive way of visualizing the best-found solution to the ESP problem (this time, found by \ikaea), in the same way that we had for the PitzDaily problem in the form of Fig. \ref{fig:curves} and Fig. \ref{fig:paraview}. Based on the statistics shown in Table \ref{tab:esp}, the best methods for the ESP problem were \tlssl, \ikaea, \samso, and \bis.

\begin{figure}[htbp]
  \centering
  \includegraphics[width=.7\linewidth]{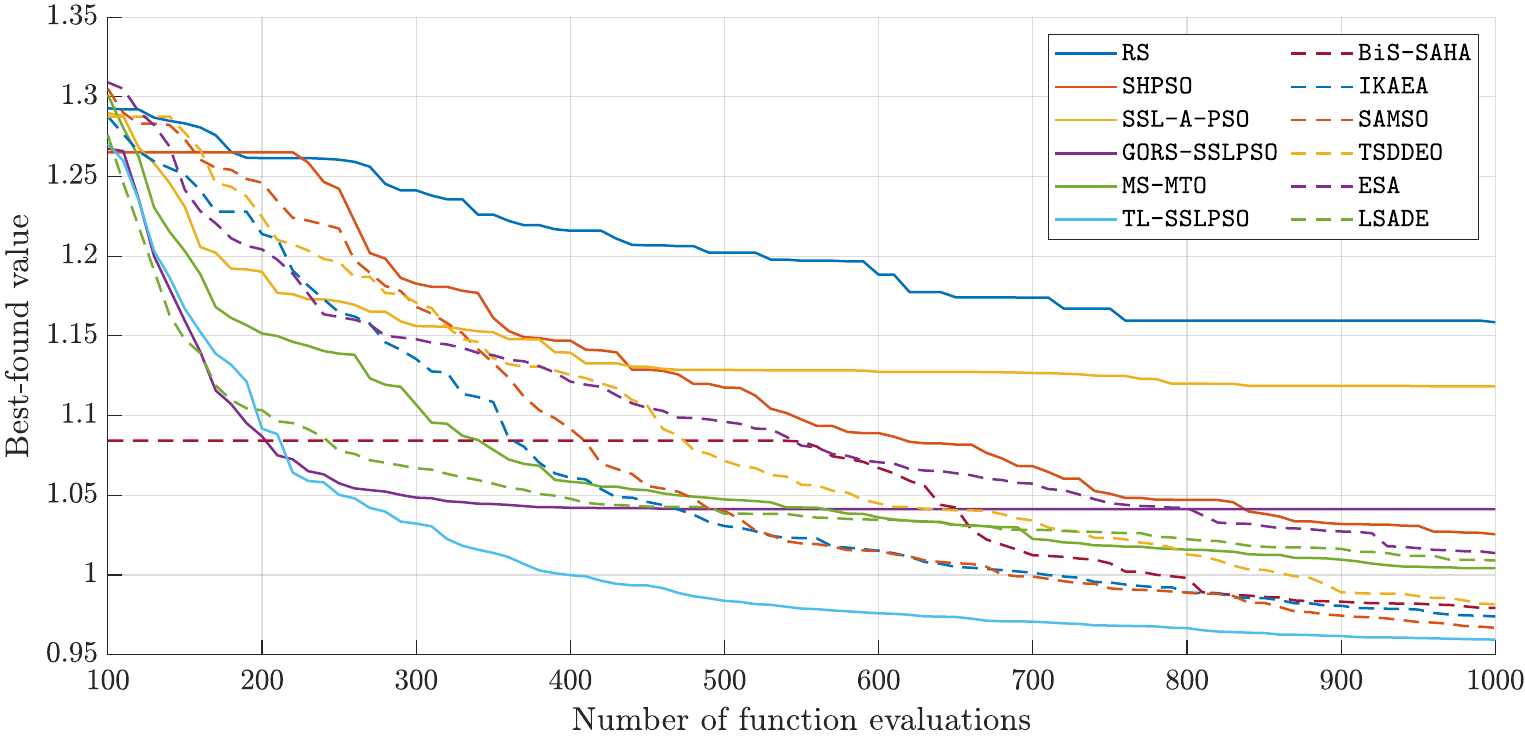}
  \caption{Convergence of the average best-found value of the selected SAEAs on the ESP problem.}
  \label{fig:conv_esp}
\end{figure}

The convergence plot, shown in Fig. \ref{fig:conv_esp}, displays some interesting patterns. There were four methods (\bis, \gors, \tlssl, and \lsade) that got relatively good solutions in a very short amount of function calls (approximately at $B=200$). However, apart from \tlssl which was the best method overall, the three other methods stagnated and got overtaken by \ikaea{} and \samso{} roughly at $B=500$. Another thing to note is that the \ssla{} method, which performed relatively well on the PitzDaily problem, was the worst of the selected SAEAs on the ESP by a large margin. Similarly to the PitzDaily case, we can find that the more recent methods dominated the older ones (apart from \tlssl).

\begin{figure}[!ht]
  \centering
  \includegraphics[width=.49\linewidth]{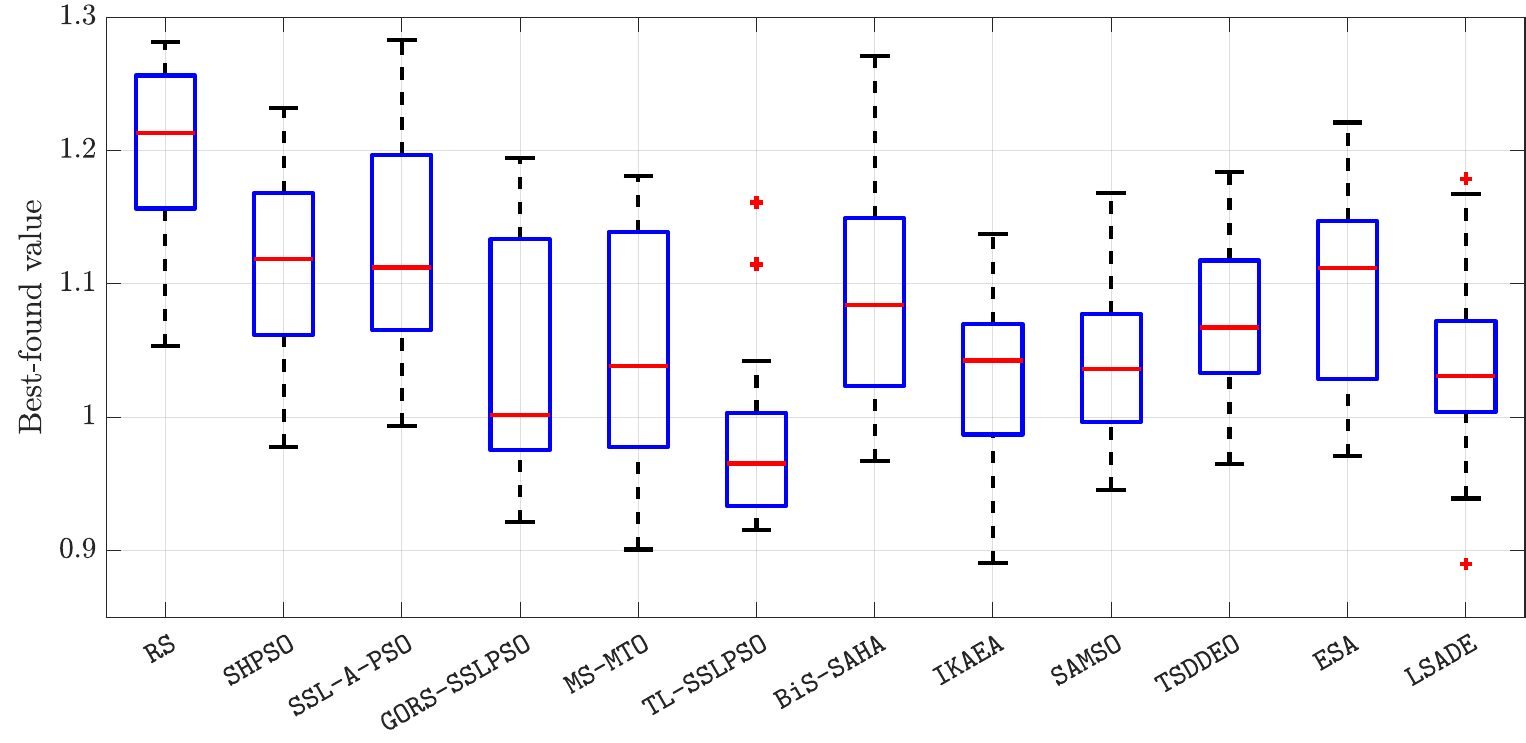}   \includegraphics[width=.49\linewidth]{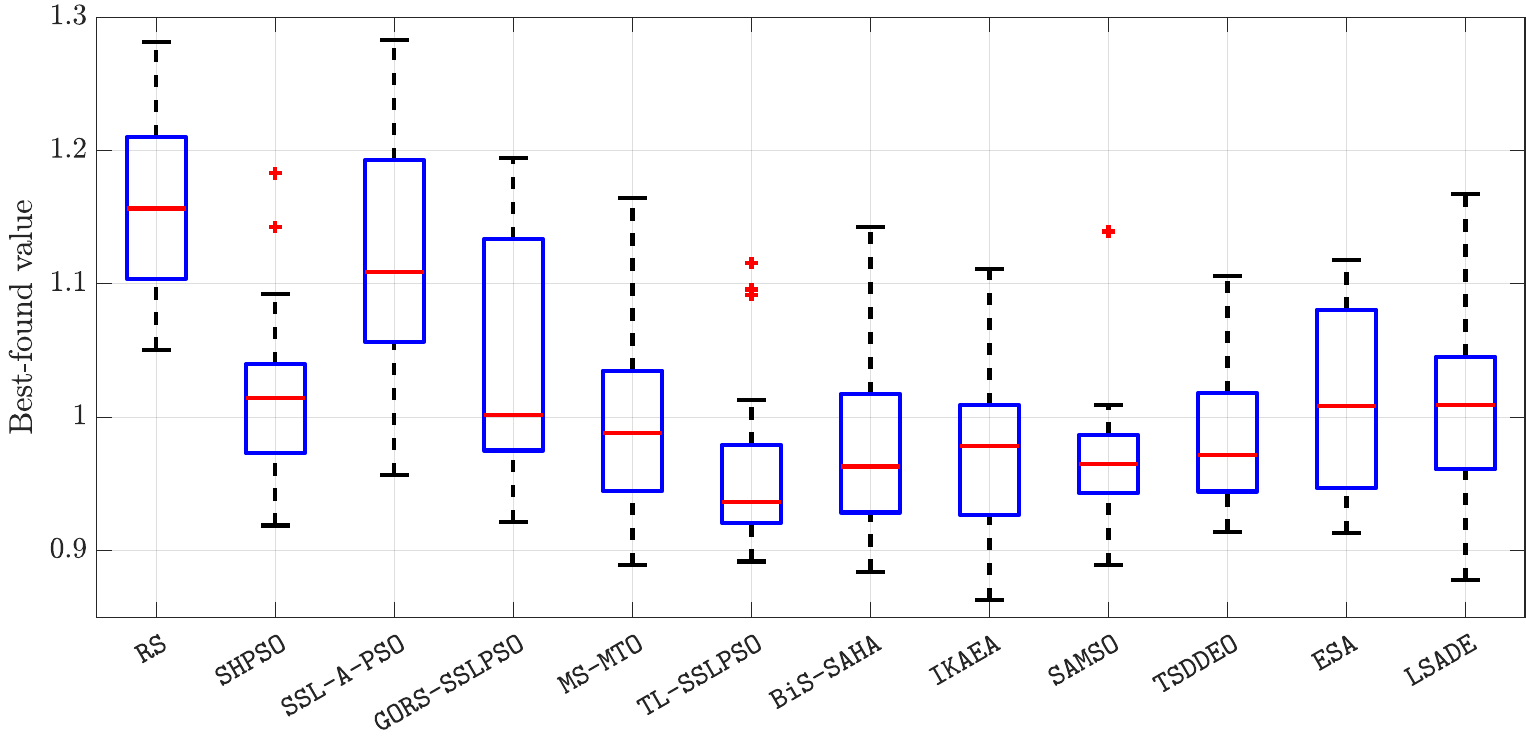}
  \caption{Boxplots of the solutions of the SAEAs on the ESP problem with $B=500$ (top) and $B=1000$ (bottom).}
  \label{fig:box_esp}
\end{figure}

The boxplots of the solutions found in the 24 runs with the two budgets ($B = 500$ and $B = 1{,}000$) are shown in Fig. \ref{fig:box_esp}. We can observe that for the ESP problem many methods, such as \lsade, \ikaea, \msmto, \gors, and \ssla, had a problem of premature convergence (not being able to improve upon a bad solution found at the $B=500$ budget). This may be explained by the difficulty these continuous SAEAs face on the discrete problems. Interestingly, the two methods with the most robust performance at the $B=1{,}000$ budget, \tlssl{} and \samso, were not the ones that were able to find the overall best solutions (these were \ikaea, \lsade, and \bis). Also, in this case, the improvement of \tsdd{} over its base method (\shpso) was substantial.

\begin{table}[]
\caption{Statistics of the 24 runs of the selected SAEAs on the ESP model. The best three methods in each category are highlighted in bold.}
\label{tab:esp}
\centering
\resizebox{0.7\linewidth}{!}{
\begin{tabular}{lcccccr} \hline 
Method      & Min      & Mean     & Median   & Max      & Std      & \multicolumn{1}{c}{Rank} \\ \hline 
\rs          & 1.05E+00 & 1.16E+00 & 1.16E+00 & 1.28E+00 & 6.46E-02 & 10.92 \\
\shpso       & 9.19E-01 & 1.02E+00 & 1.01E+00 & 1.18E+00 & 6.02E-02 & 6.92  \\
\ssla   & 9.56E-01 & 1.12E+00 & 1.11E+00 & 1.28E+00 & 9.25E-02 & 9.92  \\
\gors & 9.21E-01 & 1.04E+00 & 1.00E+00 & 1.19E+00 & 8.41E-02 & 7.50  \\
\msmto      & 8.89E-01 & 1.00E+00 & 9.88E-01 & 1.16E+00 & 8.15E-02 & 6.04  \\
\tlssl   & 8.92E-01 & \textbf{9.59E-01} & \textbf{9.37E-01} & \textbf{1.12E+00} & 6.27E-02 & \textbf{3.83}  \\
\bis    & \textbf{8.84E-01} & 9.79E-01 & \textbf{9.63E-01} & 1.14E+00 & 6.54E-02 & 4.96  \\
\ikaea       & \textbf{8.63E-01} & \textbf{9.74E-01} & 9.78E-01 & \textbf{1.11E+00} & 6.17E-02 & \textbf{4.75}  \\
\samso       & 8.89E-01 & \textbf{9.66E-01} & \textbf{9.65E-01} & 1.14E+00 & 5.04E-02 & \textbf{4.63}  \\
\tsdd     & 9.14E-01 & 9.82E-01 & 9.72E-01 & \textbf{1.11E+00} & 4.85E-02 & 5.58  \\
\esa         & 9.13E-01 & 1.01E+00 & 1.01E+00 & 1.12E+00 & 7.38E-02 & 6.46  \\
\lsade       & \textbf{8.78E-01} & 1.01E+00 & 1.01E+00 & 1.17E+00 & 6.63E-02 & 6.50 \\ \hline 
\end{tabular}
}
\end{table}

\begin{table}[!ht]
\caption{Statistical analysis of the comparison of the selected SAEAs on the ESP problem.}
\label{tab:esp_p}
\centering
\resizebox{0.7\linewidth}{!}{
\begin{tabular}{lcc|lcc}\hline 
            \multicolumn{3}{c|}{$B = 500$}  &     \multicolumn{3}{c}{$B = 1{,}000$}        \\ 
\tlssl{} vs.    & $p$        & $p^*$      & \tlssl{} vs.    & $p$        & $p^*$      \\  \hline 
\shpso       & 4.97E-05 & 4.47E-04 & \shpso       & 1.37E-03 & 1.24E-02 \\
\ssla   & 3.88E-05 & 3.88E-04 & \ssla   & 3.88E-05 & 3.88E-04 \\
\gors & 1.29E-02 & 2.80E-02 & \gors & 1.84E-03 & 1.48E-02 \\
\msmto      & 9.32E-03 & 2.80E-02 & \msmto      & 7.65E-02 & 3.82E-01\xmark \\
\bis    & 1.45E-04 & 1.01E-03 & \bis    & 1.99E-01 & 5.96E-01\xmark \\
\ikaea       & 1.29E-02 & 2.59E-02 & \ikaea       & 4.58E-01 & 6.35E-01\xmark \\
\samso       & 4.68E-03 & 2.34E-02 & \samso       & 3.17E-01 & 6.35E-01\xmark \\
\tsdd     & 9.19E-04 & 5.51E-03 & \tsdd     & 8.14E-02 & 3.82E-01\xmark \\
\esa         & 9.07E-05 & 7.25E-04 & \esa         & 2.78E-02 & 1.67E-01\xmark \\
\lsade       & 6.64E-03 & 2.66E-02 & \lsade       & 1.64E-02 & 1.15E-01\xmark \\ \hline 
\end{tabular}
}
\flushleft
$p$: p-value computed by the Wilcoxon text \\
$p^*$: p-value corrected with the Holm-Bonferroni procedure \\
\xmark: statistical differences do not exist with significance level $\alpha = 0.05$
\end{table}

As with the PitzDaily problem, we also used the Friedman test to find if significant differences are present among all the algorithms on the $B = 500$ and $B = 1{,}000$ budgets. The p-values for this test were 7.51E-15 (and 6.06E-09 when
omitting RS) for$B = 500$ and 9.13E-15 (and 3.75-08 when omitting RS) for $B = 1{,}000$. All these p-values values were
much lower than the confidence level $\alpha = 0.05$, so we can state that there are statistically significant differences
between the selected SAEAs on the ESP problem. The results of the Wilcoxon signed-rank test between the method with the lowest rank (in this case \tlssl) and the remaining methods are presented in Table \ref{tab:esp_p}. These results show the dominance of \tlssl{} on the $B=500$ budget. However, there was no statistically significant difference found between \tlssl{} and seven other methods on the $B=1{,}000$ budget. 

\subsection{Aggregate Results}
Combining the results of the PitzDaily and ESP problems, we can find the following patterns in the performance of the selected SAEAs. On very low budgets (approximately $B=200$), the \lsade{} and \gors{} algorithms were among the best methods in both cases. However, after this budget, \gors{} seemed to stagnate, while \lsade{} was able to find further improvements. On the $B=500$ budget, the situation was much more nuanced. Here, \lsade, \esa, \tsdd, and \tlssl{} were the best methods for the PitzDaily problem, while \tlssl{} dominated the ESP problem. Moreover, both \esa{} and \tsdd{} had relatively poor performance on the ESP for the $B=500$ budget. On the $B=1{,}000$ budget, the performance of many of the selected SAEAs relatively equalized, and their convergence seemed to have mostly plateaued. For both PitzDaily and ESP problems, \lsade, \esa, \tsdd, and \ikaea{} were tied for the spot of the best-performing method (based on the Wilcoxon test). The common features of these methods are their age (as they were all published after 2021), and the fact that they use DE as one of the mechanisms for optimization.

\section{Conclusions}\label{s6}
Comparing SAEAs on real-world problems instead of artificially created ones is still extremely rare. In this paper, we performed a computational comparison of eleven recently published state-of-the-art single-objective SAEAs on two CFD problems. The aggregate results of the computational comparisons showed that the overall best-performing methods (considering the quality and robustness of the obtained solution as well as their convergence properties) were \lsade, \esa, \tsdd, and \ikaea. All four of these methods were among the most recent ones (published after 2021), used DE as one of the mechanisms for optimization, and were published in the most high-profile journals (IEEE Transactions on Cybernetics, IEEE Transactions on Evolutionary Computation, and Information Sciences). We hope that these findings will help researchers faced with solving expensive optimization problems in the selection of appropriate methods. We also hope that the authors developing new SAEAs will use the results presented in this paper (and data and code available in the Zenodo repository \cite{zenodolink}) for benchmarking.

In further research, we will focus on comparisons of SAEAs with the DIRECT-type approaches \cite{jones2021direct}, which are also popular for solving computationally expensive problems. Another line of research lies in finding other good real-world applications, such as in hospital planning \cite{bartz2020hospital}, inverse heat transfer \cite{kuudela2024assessment} or robotics \cite{kuudela2023collection,juvrivcek2023evolutionary}, that can be used for benchmarking SAEAs. Lastly, benchmarking multiobjective SAEAs (and finding appropriate real-world benchmark sets) is also one of the possible research directions.

%
%
%
\bibliographystyle{splncs04}
\bibliography{library}

\end{document}